\documentclass[twocolumn]{article}

\usepackage{cite}
\usepackage{amsmath,amssymb,amsfonts}
\usepackage{algorithmic}
\usepackage{graphicx}
\usepackage{textcomp}

\usepackage{adjustbox}
\usepackage{float}
\usepackage{array}
\usepackage{pifont}  % for \Checkmark and \XSolid
\usepackage{subcaption}
\usepackage{subfiles}
\usepackage{booktabs}
\usepackage{tabularx}
\usepackage{bbding}
\usepackage{enumitem}
\usepackage[numbers,sort&compress]{natbib}
\usepackage{xcolor, colortbl}
\definecolor{highlightyellow}{RGB}{255, 255, 102}
\usepackage{soul}
\usepackage[utf8]{inputenc}
\usepackage{longtable} % For long tables spanning pages
\usepackage{makecell} % For line breaks in cells

\newcommand{\CPHSAlgo}{Local}
\usepackage{soul}

\DeclareRobustCommand{\rev}[1]{#1}
\begin{document}
\title{Federated Learning in Offline and Online EMG Decoding: A Privacy and Performance Perspective}
%\author{Kai Malcolm, Momona Yamagami}
%\thanks{This paragraph of the first footnote will contain the date on which you submitted your paper for review. It will also contain support information, including sponsor and financial support acknowledgment.}

\author{
    Kai Malcolm, C\'esar A. Uribe, Momona Yamagami \\
    \textit{Department of Electrical and Computer Engineering, Rice University, Houston, TX, USA}\\
    %\thanks{\textsuperscript{1}Kai Malcolm is with XYZ University (email: kai@example.edu).}
    %\thanks{\textsuperscript{2}Momona Yamagami is with ABC Institute (email: momona@example.org).}
    \thanks{This work was partially supported by an unrestricted gift from Meta Reality Labs. C\'esar A. Uribe is partially funded by the National Science Foundation CISE Awards \#2213568 and \#2443064, and a Google Research Award.}
}

%\author{
%  Kai Malcolm, C\'esar A. Uribe, Momona Yamagami\\
%  \textit{Department of Electrical and Computer Engineering, Rice University, Houston, TX, USA}\\
%  \thanks{This work was partially  supported by an unrestricted gift from Meta Reality Labs.}
%}

\date{} % This will suppress the date
\maketitle

\begin{abstract}
Neural interfaces offer a pathway to intuitive, high-bandwidth interaction, but the sensitive nature of neural data creates significant privacy hurdles for large-scale model training. Federated learning (FL) has emerged as a promising privacy-preserving solution, yet its efficacy in real-time, online neural interfaces remains unexplored.
In this study, we 1) propose a conceptual framework for applying FL to the distinct constraints of neural interface application and 2) provide a systematic evaluation of FL-based neural decoding using high-dimensional electromyography (EMG) across both offline simulations and a real-time, online user study. 
While offline results suggest that FL can simultaneously enhance performance and privacy, our online experiments reveal a more complex landscape. 
We found that standard FL assumptions struggle to translate to real-time, sequential interactions with human-decoder co-adaptation. Our results show that while FL retains privacy advantages, it introduces performance tensions not predicted by offline simulations.
These findings identify a critical gap in current FL methodologies and highlight the need for specialized algorithms designed to navigate the unique co-adaptive dynamics of sequential-user neural decoding.
\end{abstract}


\section{Introduction}
\label{sec:introduction}

%\IEEEPARstart{N}{eural}
Neural interfaces enable human-centric, accessible, and high-bandwidth inputs~\cite{wheelchair_gestures, MomonaDisUBGestures, Shenoy_bimanual_BCI, Shenoy_speech_BCI}.
Machine learning models (i.e., \textit{decoders}) play a crucial role in this process, translating high-dimensional neural signals into device inputs \cite{IMU_HAR, CV_GestureCustomization, CTRL_Labs, TNSRE_AttentionEEGMotorImageryDecoding, TNSRE_EMGAttention}.
% However, their efficacy depends on large quantities of biosignal data and burdensome calibration procedures. 
% This data is often scarce due to its task-specific, user-dependent nature and inherent privacy risks \cite{Priv_MomonaPT, Priv_MorrisAIAccess}.
However, high-dimensional neural signals can reveal sensitive attributes such as identity~\cite{Priv_EEGPersonIdentification, Priv_GaitPersonIdentification}, age~\cite{Priv_ECGAgeSexEstimation}, and sex~\cite{Priv_ECGAgeSexEstimation} during decoder training.
This makes individuals hesitant about contributing data and institutions cautious about open-sourcing datasets~\cite{Priv_HiddenInfoInBios, Priv_KeyloggingSideAttack, Priv_BiosensingInPublic}.
This creates a fundamental tension between the data needs for training decoders and the desire for user privacy~\cite{PPP_BigData, PPP_Survey, PPP_AttentionEcon, PPP_Smartphone, PPP_LocAwareMark, PPP_transparency, PPP_OG06}.
Federated learning (FL) offers a solution by enabling distributed training in which users train models locally and share only model updates rather than their raw neural data, eliminating the need to transmit or store sensitive neural signals centrally.

\rev{Despite its potential, applying FL to co-adaptive neural interfaces remains a significant and unexplored frontier. While FL has been extensively studied in simulated or offline settings \cite{FLwBioS_EpilepSeizPred, FLwBioS_SleepStagingEEG, FLwBioS_MitigateThreatsEEG, FLwBioS_EvoEEG, TNSRE_FL_MotorImageryClsBCI, TNSRE_FL_CerebellarAtaxia}, its application to online neural interfaces \cite{Orsborn_CLDA, CPHS, CL_CoAdaptation_DynamicMyoelectricInterface, NeuralFeedback_BMI, CL_FPGA_Spikesorting, CL_Ctrl_FES, CL_DBS} introduces unique challenges that arise when federated optimization is integrated into a real-time, online training environment. Unlike traditional FL research that focuses on aggregation across static, pre-recorded datasets, these online interfaces often involve human-decoder co-adaptation, where the user and decoder influence each other through continuous interaction \cite{CPHS, CL_CoAdaptation_DynamicMyoelectricInterface}. This dynamic introduces a critical gap: co-adaptive behaviors and real-time latency constraints may destabilize FL optimization in ways that are entirely absent in offline work. For instance, FL hyperparameters optimized in offline simulation may prove infeasible during real-time user interactions.

To address this gap, we provide the first empirical evaluation of FL in a live user study, moving beyond offline simulation to identify where standard FL assumptions break down in online training environments. Beyond performance, we explore a critical re-identification risk, demonstrating that subject identity can be recovered directly from persistent decoder weights even in the absence of raw neural data.

To facilitate future research in this nascent application, we propose a high-level framework that characterizes the inherent properties and trade-offs of different FL algorithm categories. We evaluate this framework by adapting Per-FedAvg for sequential training and comparing high-dimensional EMG decoding across offline (i.e., simulated) and online (i.e, real-time) scenarios. Our results highlight how the transition from offline simulation to online interaction alters the relationship between personalization and privacy, providing a starting point for algorithms specifically tuned for closed-loop decoder adaptation dynamics.
}

\section{Federated Learning Background}

\subsection{Privacy For Biosignals}

\rev{
A foundational challenge in data privacy is mitigating linkability and identifiability, two of the seven core threats in the LINDDUN privacy threat modeling framework \cite{Priv_LINDDUN}. Linkability is an adversary's ability to distinguish whether items of interest (e.g., medical records, model weights, etc.) originate from the same source (e.g., a user). Linkability often serves as a precursor to identifiability, or the ability to uniquely identify a subject within a system \cite{Priv_LINDDUN, Priv_KAnon}. 
This progression from linkability to identifiability was famously demonstrated by Sweeney \cite{Priv_KAnon}, who showed that even when direct identifiers (e.g., names, SSNs) are removed, unique patterns in seemingly anonymized metadata (i.e., birth date, gender, and ZIP code) can be used to uniquely re-identify individuals by linking partially overlapping datasets. In the context of biosignals, this suggests that the privacy of a system depends not only on the protection of raw data, but also on the uniqueness of the metadata (e.g., summary statistics, model weights) that are stored or transmitted.

High-dimensional non-invasive neural signals, including electroencephalography (EEG), electrocardiogram (ECG), and electromyography (EMG), are increasingly recognized for encoding unintended sensitive attributes that machine learning models can extract \cite{Priv_HiddenInfoInBios, Priv_BiosPatientID}. Prior research has demonstrated that from these signals, model can estimate demographic traits like sex and age \cite{Priv_ECGAgeSexEstimation}, detect underlying medical conditions such as depression \cite{Priv_BehavioralPrivRisks}, and uniquely identify individuals \cite{Priv_EEGPersonIdentification, Priv_SecretFromEMG, Priv_EMGFeatDis, Priv_EMGPersIdentModels}. 
EMG data presents a significant re-identification risk because it encodes distinctive motor behaviors inherent to the individual. While these biometric markers enable secure applications like device authentication and pairing \cite{Priv_SecretFromEMG, Priv_EMGFeatDis}, they can likewise be exploited for record linkage or impersonation attacks. Recent studies underscore this threat, demonstrating that deep learning models can perform re-identification or impersonate users directly from surface EMG data with high accuracy \cite{Priv_EMGFeatDis, Priv_EMGPersIdentModels}. Collectively, these findings confirm that EMG signals contain persistent physiological signatures that are uniquely and reliably linked to a specific subject.

Beyond identity disclosure, EMG signals also facilitate more advanced inference attacks. Because forearm EMG captures fine motor mechanics of the hand, it also acts as a biological keylogger \cite{Priv_KeyloggingSideAttack}, with recent work reconstructing typed passwords with over 91\% accuracy \cite{Priv_EMG2PWD}. 
Given that participants frequently express concern regarding the potential for record linkage to reveal sensitive health conditions to third parties \cite{Priv_MomonaPT}, evaluating whether FL can obscure these user-specific EMG signatures is a critical step toward securing the next generation of wearable and neural interfaces.
}

\subsection{What is Federated Learning?}

Federated learning (FL) is a type of \textit{distributed learning}, where clients collaboratively train their models while keeping their raw data local (i.e., on device). 
Specifically, a set of decentralized clients (e.g., users, nodes, etc.) communicates with a coordinating server to optimize a shared global model.
FL strikes a balance between \textit{centralized learning} (high performance but high privacy risk due to data aggregation on a central server) and \textit{local learning} (low privacy risk but limited performance due to small datasets) \cite{tang2024distributed}.
We provide a brief overview of FL for readers unfamiliar with FL. 
Knowledgeable readers may want to skip this section and return as needed.
Refer to McMahan et al. \cite{FedAvg} or Fallah et al. \cite{Per-FedAvg} for an in-depth overview of FL.

The canonical FL formulation, FedAvg \cite{FedAvg}, optimizes a global model $w$ by minimizing the cost functions of $K$ users. 
For FedAvg, the global cost function $F(w)$ is the sum of all user $i$ cost functions $f_i(w)$: 
\begin{equation} \label{eqn:canonical_fl}
     \min_{w \in \mathbb{R}^d} F(w) = \frac{1}{K} \sum^K_{i=1} f_i(w)
\end{equation}
which optimizes for a model that performs well on average for all users.
Variants such as FedProx~\cite{FedProx} and SCAFFOLD~\cite{SCAFFOLD} add regularization terms to reduce client drift, ensuring that local models stay close to the global one. These extensions aim to improve convergence under non-Independent and Identically Distributed (non-IID) conditions, which are common in biosignals \cite{Amy_CLDA_SkillfulNeuroProsCtrl, EMGHetero_EMGPatterns, EMGHetero_MultiSubj, EMGHetero_Norm}.

In neural interface applications, it is often desirable to personalize decoders to individual users.
Personalization can be incorporated into FL by shifting the optimization focus from a single shared model to user-specific models, optimizing for each user’s individual performance while retaining some shared learning or initialization via the global model.
One popular approach is Per-FedAvg \cite{Per-FedAvg}, which leverages Model-Agnostic Meta-Learning (MAML) to find a global initialization that is highly adaptable. Rather than optimizing a static global model $w$ directly (as in Equation \ref{eqn:canonical_fl}), the objective is to find a set of parameters that performs well after a small number of local gradient updates:
\begin{equation} \label{eqn:MAML}
\min_{w \in \mathbb{R}^d} F(w) = \frac{1}{K} \sum^K_{i=1} f_i(w - \alpha \nabla f_i(w))
\end{equation}
By minimizing this meta-loss, Per-FedAvg ensures that the shared global model serves as an effective prior, allowing individual users to rapidly personalize their models with minimal local computation.

\section{FL for Privacy-Preserving Neural Interfaces}
Applying federated learning (FL) to neural interfaces presents unique challenges, particularly due to the myriad of available FL algorithms and the complex demands of real-time neural systems. 
In this section, we introduce a conceptual framework to help researchers reason about FL in offline and online neural decoding and other biosignal-based scenarios.
Whereas existing frameworks (e.g., Flower \cite{flower}) look to provide the code for simulating FL algorithms, our framework focuses on guiding researchers' choice of FL algorithm for their application.
Rather than recommending a single specific algorithm, our goal is to highlight two key attributes that influence the applicability and performance of FL in neural interface studies: (1) personalized cost and (2) user availability. 
Understanding these attributes can help practitioners identify potential pitfalls and necessary algorithmic adjustments when implementing FL in privacy-sensitive neural interface applications (see Table~\ref{tab:FL_algos_wSUS}).

\subsection{Attribute 1: Personalized Cost}

FL algorithms are broadly categorized into traditional (TFL) and personalized (PFL) based on their optimization objectives. The primary driver for this selection is the data heterogeneity inherent in the problem, which dictates whether a single global model can sufficiently represent the diverse data distributions across all users. If the variance across users is high, a traditional global model may incur a high personalized cost, leading to significant performance degradation for individual participants compared to a model tailored to their specific data distributions. In practice, this heterogeneity is rarely explicitly measured; instead, researchers often rely on domain expertise to judge dataset heterogeneity and whether personalization is necessary.

\subsubsection{Traditional Federated Learning (TFL)}
TFL algorithms, such as FedAvg \cite{FedAvg}, focus on training a single, shared global model that performs well for all users. 
When all users sample from the same distribution (i.e., data across users are relatively homogeneous), the global model benefits from a large multi-user dataset, leading to improved performance. 
In comparison, training a local (i.e., non-federated) model for each user is limited to the smaller single-user dataset, which can hinder performance and generalization, and each new user must train their local model from scratch.
In the presence of substantial data heterogeneity, TFL often suffers from slow or unstable convergence and degraded performance~\cite{APFL}. 
Although some TFL variants attempt to accommodate heterogeneity~\cite{heterogeneous_fl_sota, heterogeneous_fl_learn_from_others}, the FL research community has increasingly shifted toward PFL approaches.

\subsubsection{Personalized Federated Learning (PFL)}
PFL algorithms can be grouped into two categories: those addressing device heterogeneity (e.g., network reliability and hardware capabilities) and those addressing data heterogeneity. In this work, we concentrate on the latter.
PFL optimizes a set of user-specific models tailored to each user's data distribution, often using the global model as a prior or initialization to improve generalization and stability. These approaches are made to be more robust to distributional variation but are often more complex to tune and slower to converge~\cite{PersAFL}. 
If dataset heterogeneity is unknown, personalized models may offer a more robust solution, trading faster convergence for better performance.
A growing number of PFL algorithms address diverse personalization scenarios, including Per-FedAvg~\cite{Per-FedAvg}, pFedMe~\cite{pFedMe}, APFL~\cite{APFL}, FedALA~\cite{FedALA}, Fedcp~\cite{Fedcp}, FedDBE~\cite{FedDBE}, and GPFL~\cite{GPFL}, with many works benchmarked on standardized image datasets such as CIFAR-100 \cite{CIFAR100}.

\begin{table*}[!t]
\centering
\renewcommand{\arraystretch}{1.75}
\caption{
Categorization of FL algorithms based on relevant dimensions (see Zhang \cite{tsingz0} for a large but non-comprehensive list, with PyTorch code). 
}
\label{tab:FL_algos_wSUS}
\begin{tabularx}{\textwidth}{l l X}
\hline
% \shortstack allows for manual line breaks within a single cell
\textbf{\shortstack[l]{Dimension 1: \\ Personalized Cost}} & 
\textbf{\shortstack[l]{Dimension 2: \\ User Availability}} & 
\textbf{Example Algorithms} \\
\hline
Traditional  & Synchronous  & FedAvg \cite{FedAvg}, FedProx \cite{FedProx}, FedDyn \cite{FedDyn}, FedNTD \cite{FedNTD} \\
Personalized & Synchronous  & Per-FedAvg \cite{Per-FedAvg}, pFedMe \cite{pFedMe}, FedALA \cite{FedALA}, GPFL \cite{GPFL}, FedDBE \cite{FedDBE} \\
Traditional  & Asynchronous & FedAsync \cite{FedAsync}, FedSA \cite{FedSA}, AsyncDrop \cite{HAsyncDrop} \\
Personalized & Asynchronous & PersA-FL-MAML \cite{PersAFL}, PersA-FL-ME \cite{PersAFL}, Hetero AsyncDrop \cite{HAsyncDrop} \\
\hline
\end{tabularx}
\end{table*}

\subsection{Attribute 2: User Availability}
User availability refers to how frequently and reliably user information can be integrated back into the network. 
This includes real-world issues such as users dropping out of the network, differences in computational resources, and variations in network connection speeds (e.g., delays). Our framework considers three modes of user availability: synchronous, asynchronous, and sequential. 

\subsubsection{Synchronous}
Synchronous FL assumes that all users are available simultaneously. This is the default assumption for most FL algorithms ~\cite{FedAvg, FedALA, FedDBE, FedDyn, FedProx, pFedMe, Per-FedAvg}.
While simple and easy to simulate, synchronous FL has limited applicability to laboratory-based, online neural interface experiments, where running multiple users simultaneously is often infeasible due to hardware or personnel constraints.
Looking towards deployment, synchronous FL also struggles with straggling users that require substantially more time: slow users delay the entire training process, and users unexpectedly dropping out of the network can halt training indefinitely~\cite{FL_straggler}.
Synchronous algorithms are often sufficient for offline simulations, where real-time latency, hardware, and personnel constraints do not apply.

\subsubsection{Asynchronous}
Asynchronous FL relaxes the requirement of simultaneous user availability.
Typically, users can query the server at their own pace, thus reducing sensitivity to stragglers and dropouts \cite{PersAFL}. 
However, this flexibility introduces new challenges. First, updates from slower users may be based on outdated global models, leading to ``stale'' updates relative to the current global model. Second, faster users can dominate the training process, introducing bias (i.e., over-representation) into the global model. 
More recent algorithms address these issues through fairness and weighting mechanisms~\cite{IMWUTFL_AreYouLeftOut_Fair, IMWUTFL_BiasMitigation}.
Asynchronous methods offer a more robust alternative to synchronous FL, particularly for real-world deployment. However, asynchronous approaches still assume a small group of concurrently active users. 
This ongoing stream of diverse updates is key, as it serves as implicit regularization, helping prevent overfitting to any single user.

\subsubsection{Sequential}

We introduce a novel learning scenario we term \textit{sequential federated learning}, defined by two constraints: (1) only one user is active at a time, and (2) each user appears exactly once, fully completing their training during this window. 
This regime is highly relevant for online decoding experiments, such as EMG or brain-computer interface (BCI) studies \cite{ELEC677_EMGNeuralMachineInterface, ELEC677_UnsupTrackingLatentManifolds, ELEC677_FuncReorgBoMI, ELEC677_GaitResponseAnkleExo, ELEC677_PersExo}, where participants contribute data in isolation, and repeated access to a given user is uncommon outside of larger-scale longitudinal studies. 
Despite its practical importance, no existing FL algorithms are designed for this scenario.

Sequential FL breaks a core assumption of FL that updates from multiple users can be aggregated via averaging each training round. 
Since only one client is active per round, averaging-based model aggregation becomes model replacement: the global model is overwritten with the (sole) active user's uploaded model. 
This diminishes the stabilizing effect of averaging over users, increasing the likelihood of overfitting to the active user. 
Furthermore, if a user's model drifts far from the current global model, this can exacerbate client drift and hinder convergence. 
When users are seen only once, mitigating overfitting to individual users becomes critical. 

Although no prior work addresses this exact setting, a few related paradigms exist. 
For instance, \textit{cyclical FL}~\cite{CyclicFL, CyclicFL_Medical} also assumes one user trains at a time but relaxes the single-pass constraint by returning to previous users in a cyclical fashion. 
This approach is often applied in the institutional context, with a small number of persistent institutions that pass the model weights around in a continuous cycle, thus violating the second constraint of our sequential FL definition.
In summary, our formulation of sequential FL introduces unique challenges not addressed by prior work, motivating the development of new methods for real-world online neural applications.

\section{Methods}
\label{sec:methods}

The goal of our empirical evaluation is to analyze how decoder performance and privacy risk vary not only across learning strategies (i.e., local learning versus FL) but also across two key settings for high-dimensional neural interface control: offline and online control. By running matched experiments under both conditions, we assess how learning paradigms generalize and adapt in offline versus online decoding scenarios.
In particular, in our online user study, we analyze closed-loop decoder adaptation (CLDA), in which the decoder’s parameters are updated continuously during interaction to accommodate changes in the user’s neural (myoelectric) signals. This process creates a dynamic, co-adaptive relationship in which the user modifies their motor strategy in response to the decoder's performance, and simultaneously, the FL algorithm optimizes the model weights to track the user's shifting physiological patterns.
%Given the success shown with online adaptation~\cite{Orsborn_CLDA}, we expect online convergence to be influenced by this human-interface co-adaptation process~\cite{madduri2024predicting}.

\subsection{Task and Adaptive Decoder Description}

\begin{figure}[h]
    \centering
    \includegraphics[width=0.9\columnwidth]{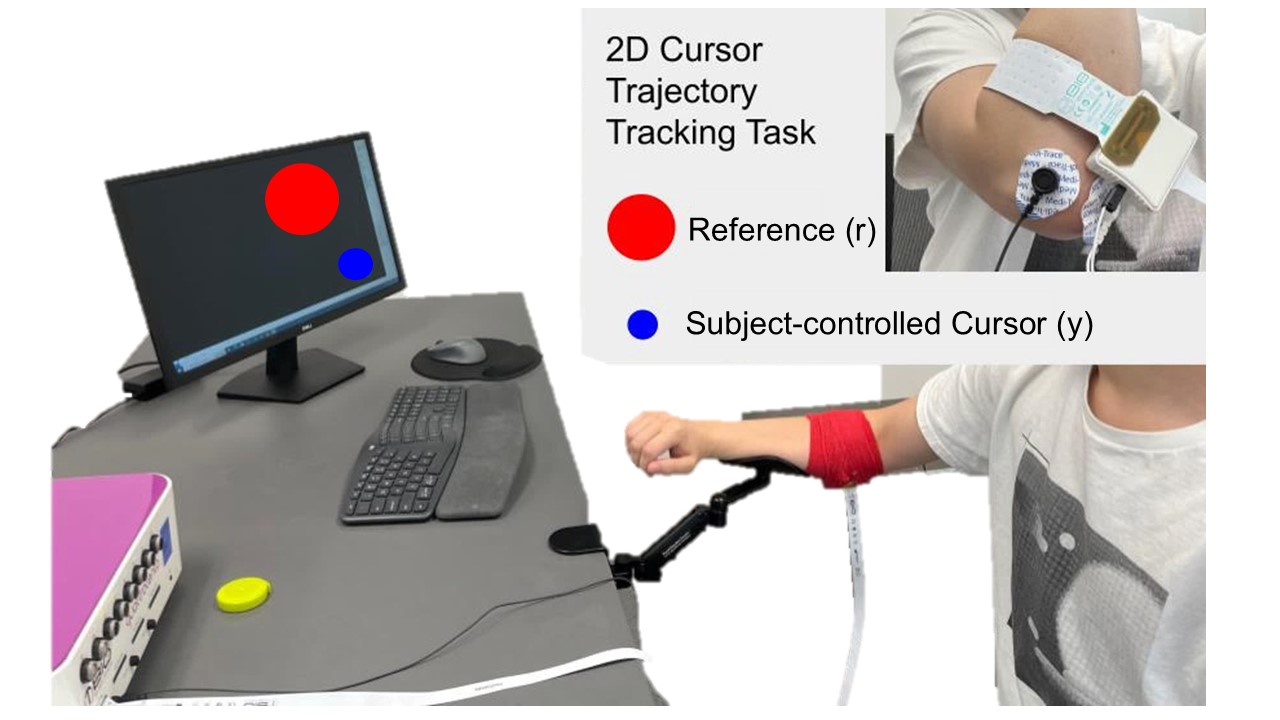}
    \caption{Experimental setup. The subject controls a blue cursor using their forearm muscles, with activity recorded from a 64-channel Quattrocento EMG sensor. The subject's goal is to track the red target by minimizing the distance between the blue cursor and the red reference trajectory.}
    \label{fig:setup}
    %\Description{A photo of a man sitting at a desktop computer with their arm on an armrest. The computer has a 2D trajectory tracking task consisting of a large red circle and a smaller blue circle on a black screen, and the man has wired surface EMG electrodes just below his elbow secured with red athletic tape. There is a closeup picture in the top right corner of the figure showing the EMG electrodes without the red athletic tape obscuring them, and a legend explaining that the larger red circle on the computer screen is the target and the smaller blue circle is the subject-controlled cursor.}
\end{figure}

The offline study is a secondary data analysis of the data collected in Madduri et al. \cite{CPHS}, and the online study experiment adapts the same task of Madduri et al. \cite{CPHS} to facilitate direct comparison.
Briefly, subjects were asked to control a cursor $y$ in the horizontal and vertical directions on the screen to track a reference $r$ using an EMG interface placed on their forearm (Fig.~\ref{fig:setup}). 
The reference trajectory was the sum of two sinusoids with frequencies 0.10 and 0.25 Hz in the horizontal direction and 0.15 and 0.35 Hz in the vertical direction, each starting each trial with a random phase to prevent trajectory memorization. 

\subsubsection{EMG Data Collection and Filtering}
EMG data were collected using the Quattrocento system (Bioelettronica, Italy), measuring the dominant hand's forearm muscle activity with a 64-channel electrode array (5×13 electrode layout). All subjects were right-handed.
EMG data was collected at 2048 Hz and filtered via built-in analog high- and low-pass filters at 10 Hz and 130 Hz, respectively.
We then computed the EMG linear envelope following~\cite{Momona_EMG}, and applied a rolling average with a window size of 250 ms and 93\% overlap to downsample the EMG data from 2048 Hz to 60 Hz.
% https://github.com/kdmalc/labgraph-Rice/blob/b5eb55ef3def570d004ab0f805ca4090f277a982/experiment/nodes/emg_inputs.py#L184

\subsubsection{EMG-to-Cursor Decoder and Adaptation}

Detailed methods and decoder parameter analysis can be found in~\cite{CPHS}.
Briefly, linear regression was used to convert the streamed EMG data
($s\in\mathbb{R}^{64\times 1}$) into the $x$ and $y$ cursor velocity ($v=w s$, where $w \in\mathbb{R}^{2\times 64}$ are the decoder parameters and $v\in\mathbb{R}^{2\times 1}$).
The cursor velocity $v$ was numerically integrated to obtain the horizontal and vertical cursor positions. 
The decoder parameters $w$ were updated in roughly 20-second batches by minimizing the sum of the squared difference between the velocity error and the magnitudes of the decoder parameters. The velocity error is defined as the difference between the decoder's predicted velocity and the optimal gap-closing velocity between the reference $r$ and the user's cursor $y$ velocities:
\begin{equation}\label{eq:cost_func}
    \min\limits_{w} f(w) = ||w s - \frac{\partial}{\partial t}(r-y)||_2^2 + \lambda||w||_2^2,
\end{equation}
where $\lambda_{w}$ is the penalty parameter associated with the L2 norm on the decoder parameters $w$.
All decoder updates implemented SmoothBatch~\cite{Orsborn_CLDA}, where the optimal decoder parameters are linearly combined with the old decoder parameters $w_{new} = (1-\alpha) w_{old} + \alpha w_{optimal}$, with learning rate $\alpha$.
In line with prior work~\cite{CPHS}, our pilot online studies found that a penalty parameter of $\lambda=10^{2}$ and a slow SmoothBatch learning rate of $\alpha=0.75$ performed well; hence, these were used in our offline and online studies. 
The dataset collected by Madduri et al. \cite{CPHS} was an online experiment, but did not use FL: rather, each subject trained their own local decoder using what we refer to as the \textit{\CPHSAlgo{}} algorithm. 
Each subject started with random initial decoder parameters and adapted simultaneously with the decoder during each 5-minute trial. 

\subsection{Choosing an FL Algorithm: Applying Attributes to Determine Suitability}

We used Table \ref{tab:FL_algos_wSUS} for our offline and online EMG decoding to identify possible FL algorithms using the two attributes: personalized cost and user availability. 

\subsubsection{Offline Study}

% In our offline study, we would like to implement both PFL and TFL algorithms to compare their performance.
For the first attribute, \textit{personalized cost}, we use our domain expertise to estimate the dataset's heterogeneity. 
It is well known that many types of neural signals are highly unique and heterogeneous \cite{Amy_CLDA_SkillfulNeuroProsCtrl, EMGHetero_EMGPatterns, EMGHetero_fMRI, EMGHetero_MultiSubj, EMGHetero_Norm, EMGHetero_QTRR}. Thus, for high-dimensional neural interfaces, we can assume the dataset is heterogeneous.
This would lead us to focus on PFL algorithms. However, we would like to use the offline study to investigate whether PFL or TFL algorithms perform best in our specific domain; thus, we will implement both a TFL and a PFL algorithm.
For the second attribute, \textit{user availability}, our offline study (i.e., simulation) provides full control over user availability, as we can set the constraints.
Thus, we chose to simulate the standard synchronous case, where all users are available at the same time.
Considering the results of both dimensions, we accept any synchronous TFL and PFL algorithms, corresponding to rows 1 and 2 in Table \ref{tab:FL_algos_wSUS}.
% Considering the results of both dimensions, we highlight the algorithms that meet our minimum criteria in Table \ref{tab:FL_algos_wSUS}.
%
For PFL algorithms, we have no prior on which PFL algorithm (e.g., Per-FedAvg, pFedMe, etc.) will perform best on our dataset; thus, we implement Per-FedAvg as a baseline based on its popularity in other works \cite{FedDBE, GPFL, Fedcp, PFLBridging, PFL_SharedReprs, PFL_FirstOrder}. 
Likewise for TFL algorithms, from the acceptable Table \ref{tab:FL_algos_wSUS} row 1 algorithms, we choose to implement FedAvg for our case studies, as this is the original FL algorithm and thus a good starting point and baseline for researchers new to FL. 
%
% Thus, we implement our baseline \CPHSAlgo{} algorithm (not federated, not part of the framework), FedAvg, and Per-FedAvg, which are traditional non-federated machine learning, traditional FL, and personalized FL, respectively.

\subsubsection{Online Study}

For the first attribute, \textit{personalized cost}, we assume the dataset is heterogeneous and therefore best suited to a PFL algorithm. 

For the second attribute, \textit{user availability}, when running online studies in the lab environment, we are constrained to using a single data-acquisition device, so only one participant can participate in the experiment at a time. 
This necessitates a sequential FL setup, where users train one at a time and do not return for subsequent rounds. However, no such algorithms exist.
Thus, we use our framework to determine which algorithm to start with.
Since we are not constrained by real-world deployment challenges such as stragglers or intermittent connectivity, we do not require the robustness offered by asynchronous FL methods.
Therefore, we adapt an existing synchronous FL algorithm for the sequential setting. 
This allows us to preserve algorithmic simplicity while accommodating only the necessary constraints of our online, real-time experiment.

Taking the results of both attributes into consideration, the algorithms of row 2 in Table \ref{tab:FL_algos_wSUS} satisfy these attributes.
Given the proliferation of Per-FedAvg as a baseline ~\cite{pFedMe, GPFL, DITTO, FedDBE, FedALA} and our desire to match our offline study, we chose Per-FedAvg over the other PFL algorithms.

\subsection{Adaptive Algorithm Implementation}

We implemented a baseline non-federated algorithm (\textit{\CPHSAlgo{}}) and two FL algorithms (\textit{Per-FedAvg} and \textit{FedAvg}) for the offline study and a \textit{Local} and FL algorithm (modified \textit{Per-FedAvg}) for the online study.~\footnote{The code is available at https://github.com/kdmalc/framework-for-fl-emg}

\subsubsection{Local}
We implemented the same \textit{Local} learning algorithm used in prior work~\cite{CPHS}, where Equation~\ref{eq:cost_func} is minimized using the Python package \texttt{scipy.minimize()}. 
We ran SmoothBatch \cite{Orsborn_CLDA} between the user's current decoder weights and the new optimal decoder weights to ensure that the decoder does not change too drastically for the user.
This algorithm is subject-specific, using only the data from the subject's most recently streamed update.
In both the offline and online studies, \CPHSAlgo{} fully minimized each streamed update, with each update comprising 20 seconds of data.

\subsubsection{Per-FedAvg and Sequential Modification}
\textit{Per-FedAvg} optimizes the MAML cost function (Eq. \ref{eqn:MAML}). For both the offline and online implementations, we implemented first-order Per-FedAvg, approximating the Hessian with a first-order term.

In the offline study, Per-FedAvg was run for 500 global rounds, selecting 35\% of clients per round and performing 25 gradient steps per client. 
Each client advanced to their next streamed update after completing 25 local training rounds, meaning clients progressed at different rates due to the random sampling of clients each round.
Once a client advanced to the final streamed update, they used it for the rest of the simulation.

In the online study, we ran 40 training rounds on each streamed update, with 20 seconds of data per update. We divided each update into 6 batches to achieve real-time speed.
To accommodate the sequential nature of our human-subject study, we modified the original Per-FedAvg algorithm \cite{Per-FedAvg} by replacing the standard multi-user averaging scheme with a weighted average between the active subject’s updated weights and the server's current global weights. 
In our linear implementation, this modification is mathematically equivalent to SmoothBatch regularization \cite{Orsborn_CLDA}, which constrains the local decoder to remain in proximity to the global parameters. 
Although this approach removes simultaneous cross-user averaging, the global model still serves as a shared, evolving initialization that is passed between participants, effectively linking sequential local learning optimizations through a shared global prior.

\subsubsection{FedAvg} 
%The FedAvg algorithm is the baseline algorithm for FL, coming from the paper that first introduced FL \cite{FedAvg}. 
FedAvg is implemented only for the offline study and shares all of Per-FedAvg's described hyperparameters, with the only difference being the cost function it optimizes (Eq. \ref{eqn:canonical_fl}).

%[!t]
\begin{table*}[tb]
\centering
\caption{\rev{Comparison of offline and online evaluation settings.}}
%\caption{Comparison of offline and online evaluation settings.}
\begin{subtable}{0.48\textwidth}
\centering
\renewcommand{\arraystretch}{1.2}
\caption{Offline (Synchronous, N=14)}
\begin{tabular}{c c c c}
\hline
Algorithm & Eval. Scenario & Model Init. \\
\hline
\CPHSAlgo{} \cite{CPHS} & Intra-Subj. & Random \\
\CPHSAlgo{} \cite{CPHS} & Cross-Subj. & Random \\
FedAvg \cite{FedAvg} & Intra-Subj. & Random \\
FedAvg \cite{FedAvg} & Cross-Subj. & Random \\
Per-FedAvg \cite{Per-FedAvg} & Intra-Subj. & Random \\
Per-FedAvg \cite{Per-FedAvg} & Cross-Subj. & Random \\
\hline
\end{tabular}
\end{subtable}
\hfill
\begin{subtable}{0.48\textwidth}
\centering
\renewcommand{\arraystretch}{1.2}
\caption{Online (Sequential, N=16)}
\begin{tabular}{c c c c}
\hline
Algorithm & Eval. Scenario & Model Init. \\
\hline
\CPHSAlgo{} \cite{CPHS} & Intra-Subj. & Random \\
\CPHSAlgo{} \cite{CPHS} & Intra-Subj. & Offline \\
Static & Intra-Subj. & Pretrained \\
Modified Per-FedAvg & Intra-Subj. & Random \\
Modified Per-FedAvg & Intra-Subj. & Offline \\
\hline
\end{tabular}
\end{subtable}
\end{table*}

\subsection{Outcome Measures}

Our primary outcomes for both the offline and online studies were 1) \textit{Velocity Error} and 2) \textit{Model Re-identification Risk}. 
These outcomes compare each decoder's performance and privacy risk across the tested FL algorithms.

\subsubsection{Velocity Error}
We quantified decoder performance using velocity error, which is equivalent to an unscaled version of the first term of the user cost function:
\begin{equation}\label{eq:vel_error}
    v_e = ||w s - \frac{\partial}{\partial t}(r-y)||_2^2 
\end{equation}
The last full decoder update was used to evaluate each user's test performance.

%%%%%%%%%%%%%%%%%%%%%%%%%%%%%%%%%%%%%%%%%%%%%%%%%%%
%%%%%%%%%%%%%%%%%%%%%%%%%%%%%%%%%%%%%%%%%%%%%%%%%%%
\subsubsection{Model Re-identification Risk}

\rev{In our threat model, we assume the raw EMG data remains on the user device and is never directly at risk. However, we assume an adversary may intercept the communicated model weights or access leaked model weights via a server-side breach. 
Inspired by recent works taking advantage of gradient leakage \cite{DLG, Priv_FedACDP_DLG} as well as foundational works in record linkage attacks \cite{Priv_LINDDUN, Priv_KAnon}, we quantified Model Re-identification Risk as the ability (i.e., accuracy) of an adversarial classifier to infer subject identity from trained decoder weights. 
High identification accuracy suggests that the trained decoder weights encode subject-specific information, potentially increasing vulnerability to more advanced attacks commonly conducted against deep learning models (e.g., membership inference~\cite{MemInfAttack} or model inversion~\cite{ModelInvAttack}). 
We distinguish this specific technical metric, which measures the linkability of model weights to individuals, from broader user privacy concerns, such as the explicit disclosure of sensitive health conditions to third parties \cite{Priv_MomonaPT} or model inversion attacks \cite{ModelInvAttack}.
}

\rev{We conducted our adversarial attack on the decoder weights rather than gradients for two reasons. First, the FL algorithms explored in this study (FedAvg and Per-FedAvg) both communicate model weights rather than gradients, thereby making the model weights vulnerable during distributed training. Second, while gradients are typically not stored, model weights are necessarily stored and often backed up, representing a more persistent threat model in both Local and FL settings. In particular, for models trained solely on-device (i.e., our non-federated Local models), the gradients are never extracted or communicated, thus making gradient-based attacks unrealistic. Moreover, gradients typically leak more information than weights \cite{DLG}; thus, our approach provides a conservative lower bound of the re-identification risk.}

To evaluate this privacy risk, we extracted a representative set of decoder snapshots from each user and condition. 
Specifically, we selected the final six decoder updates per user (i.e., the final six streamed updates, by which point the model should have been sufficiently trained), ensuring consistency across conditions. We flattened these decoder snapshots into vectors and assembled them into comprehensive datasets covering all subjects.
We then performed leave-one-sample-out cross-validation to quantify the adversary's ability to correctly identify subjects from their decoder weights. 
For each sample, we trained a support vector classifier (SVC) on all other samples and assessed whether the model correctly identified the held-out sample's subject identity. By aggregating these predictions over all iterations, we obtained a per-subject accuracy measure, reflecting each subject's Model Re-identification Risk.
%%%%%%%%%%%%%%%%%%%%%%%%%%%%%%%%%%%%%%%%%%%%%%%%%%%
%%%%%%%%%%%%%%%%%%%%%%%%%%%%%%%%%%%%%%%%%%%%%%%%%%%

\subsection{Offline Evaluation and Analysis}

\subsubsection{Dataset}

The offline evaluation presented here used only the data from the best-performing condition from Madduri et al. \cite{CPHS}: a slow (SmoothBatch) learning rate ($\alpha=0.75$), a low decoder parameter penalty ($10^{2}$), and a positive decoder parameter initialization. 
In total, each trial consisted of five minutes of data for each of the 14 subjects.
Because we did not simulate the subjects' changing EMG responses to adaptive decoders, we excluded the initial two updates (of 18 total) to allow initial learning and adaptation effects to stabilize. Thus, our analysis used data from update three onward (approximately 4 minutes of the 5-minute trial duration).%, resulting in roughly 17,000 time points per subject.

\subsubsection{Data Streaming and Train/Test Split}

In the data-collection experiment, each subject's data was streamed in 18 consecutive updates, each containing 20 seconds of data. 
To replicate this streaming effect in our offline study, we segmented the dataset so that the decoder trains on one update at a time. After training for a predefined number of rounds, the simulated subject advances to the next update in the sequence.

\begin{figure}[htbp]
  \centering
  %,trim={1cm 2cm 4cm 2cm},clip]
  \includegraphics[width=1\columnwidth]{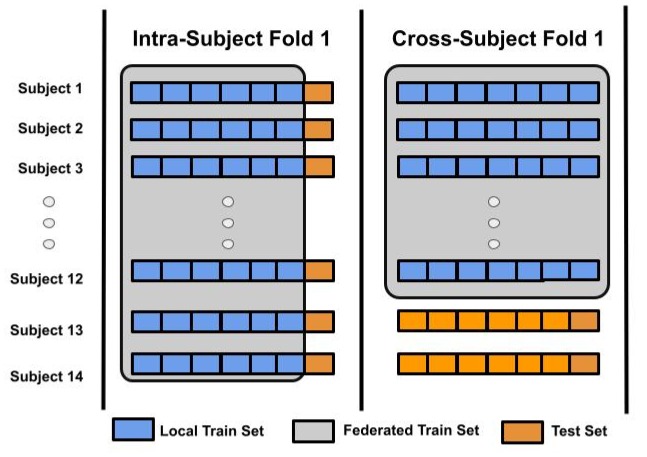}
  \caption{Offline data splits: \CPHSAlgo{} train (blue), FL train (grey), and test (orange) split for the first fold in seven-fold cross-validation ($N=14$ subjects). 
  For the \textit{Intra-Subject Scenario}, the federated decoder is trained across six training folds for all 14 subjects.
  Each subject-specific \CPHSAlgo{} decoder is trained on a single subject's six training folds and evaluated on that subject's withheld testing fold.
  For the \textit{Cross-Subject Scenario}, the federated decoder is trained on the entire dataset for Subjects 1-12, and cross-subject decoders are evaluated on Subjects 13-14. Likewise, the \CPHSAlgo{} decoders are trained on a single subject's data, and all are tested on the same withheld subjects.}
  \label{fig:IntraVsCrossScenarios}
\end{figure}

\subsubsection{Evaluation Design}

% The primary reported outcomes for all conditions were Tracking Error and Empirical Privacy Risk. 
The offline evaluation was performed with the following factors and levels:
\begin{itemize}[leftmargin=*]
    \setlength\itemsep{0em}
    \item \textit{Algorithm} (within-subject): Local, Per-FedAvg, FedAvg;
    \item \textit{Scenario} (within-subject): Intra-Subject, Cross-Subject.
\end{itemize}
We evaluated two \textit{Scenarios}: Intra-Subject and Cross-Subject. For both scenarios, we followed a k-fold cross-validation test/train split with k=7.
In the \textit{Intra-Subject Scenario}, we implemented a data split within each user, with some of their data withheld for the final outcome evaluation.
Each subject's dataset is divided into seven folds, with one fold held out as the test set during each evaluation (Fig.~\ref{fig:IntraVsCrossScenarios}, left). 
In the \textit{Cross-Subject Scenario}, we tested decoders on data from withheld subjects that were not included in the training dataset.
Here, we implemented a user-based data split by randomly assigning subjects to seven groups. For each fold, one group was used as the testing subjects and the other six groups were used as the training subjects (Fig.~\ref{fig:IntraVsCrossScenarios}, right). 

\subsubsection{Statistical Analysis}

For the offline studies, we found that the residuals from a $2\times 2$ within-subjects factorial repeated-measures ANOVA model for the dataset were highly non-normal. 
Therefore, we performed a rank transformation on the outputs~\cite{conover1981rank} and found that the residuals of the transformed ANOVA model were normally distributed (Kolmogorov-Smirnov test: $p>0.05$). 
Holm's sequential Bonferroni procedure~\cite{holm1979simple} was used to correct for multiple comparisons when performing \textit{post-hoc} tests on significant main and interaction effects.

\subsection{Online Evaluation and Analysis}

Our protocol was approved by Rice University's Institutional Review Board (IRB-FY2024-3) on August 28th, 2023; informed consent was obtained from all participants. 

\subsubsection{Real-Time Implementation}
In our online study, we used batches of 20 seconds, as pilot studies found that this was optimal for FL's stability. Thus, over the roughly five-minute trial, we had 15 updates. 
The reference tracking task, real-time EMG streaming, and decoder parameter adaptation were implemented in \texttt{Python3} using LabGraph \cite{LabGraph}. %~\cite{https://github.com/facebookresearch/labgraph}.

\subsubsection{Subject Demographics}
16 people (8 women, 8 men) participated in the study. 
Subjects were all right-handed (61\% all dominant, 39\% somewhat ambidextrous in handedness), with an average weight and standard deviation of 151.1 ± 36.8 lbs, a height of 65.9 ± 3.6 in, a forearm circumference of 10.1 ± 1.1 in, and an age of 23 ± 2 years. All subjects were daily computer users.

\subsubsection{Evaluation Design}

The online evaluation was performed with the following factors and levels:

\begin{itemize}[leftmargin=*]
    \setlength\itemsep{0em}
    \item \textit{Algorithm} (within-subject): Local, Per-FedAvg, Static;
    \item \textit{Initialization} (within-subject): Random, Simulation. 
\end{itemize}
For \textit{Algorithm}, in addition to \textit{\CPHSAlgo{}} and \textit{Per-FedAvg}, subjects also completed a ``Static'' condition, where they used their previously trained \textit{\CPHSAlgo{}} decoder, but during this trial, no further decoder training happened. This provides a baseline of how subjects learn and adapt to fixed decoders without co-adaptation.
We initialized the Static condition using subject-specific, previously trained \CPHSAlgo{} decoders, rather than a generic global decoder or one pre-trained on either offline simulations or an expert EMG user, because subjects could not gain control with these decoders during pilot testing. 

We evaluated two \textit{Initializations} for \CPHSAlgo{} and Per-FedAvg: Random and offline. 
In the \textit{Random Initialization} condition, the decoder weights were randomly generated as positive values between 0 and 1; all subjects used the same set of weights.
In the \textit{Offline Initialization} condition, all clients started from the final Per-FedAvg global decoder from our offline study. 

To minimize learning and fatigue effects, trial orders were randomized for each subject, following a Latin Square design. 

\subsubsection{Statistical Analysis}

For the online studies, we found that the residuals from a $2\times 2$ within-subjects factorial repeated-measures ANOVA model for the dataset were highly non-normal. 
Therefore, we performed a rank transformation on the outputs~\cite{conover1981rank} but found that the residuals of the transformed ANOVA model were still not normally distributed (Kolmogorov-Smirnov test: $p<0.0005$). 
Thus, we ran an aligned rank transform (ART) ANOVA to fit a nonparametric factorial repeated-measures model. 
We also performed planned paired Wilcoxon signed-rank tests comparing Static with all other conditions.
The Holm's sequential Bonferroni procedure ~\cite{holm1979simple} was used to adjust for multiple comparisons. 

\subsection{Offline and Online Decoder Analysis}

We analyzed the convergence of decoders in two ways: first, for each trial, we took the Euclidean distance between each decoder and that trial's final decoder. Second, we flattened all decoders, mean-subtracted them, and then used PCA projections to visualize them in a two-dimensional subspace.

\section{Results}

\subsection{Offline Result: FL Algorithms Outperform Traditional Approaches}
\begin{figure}[htbp]
  \centering
  \includegraphics[width=0.45\textwidth]{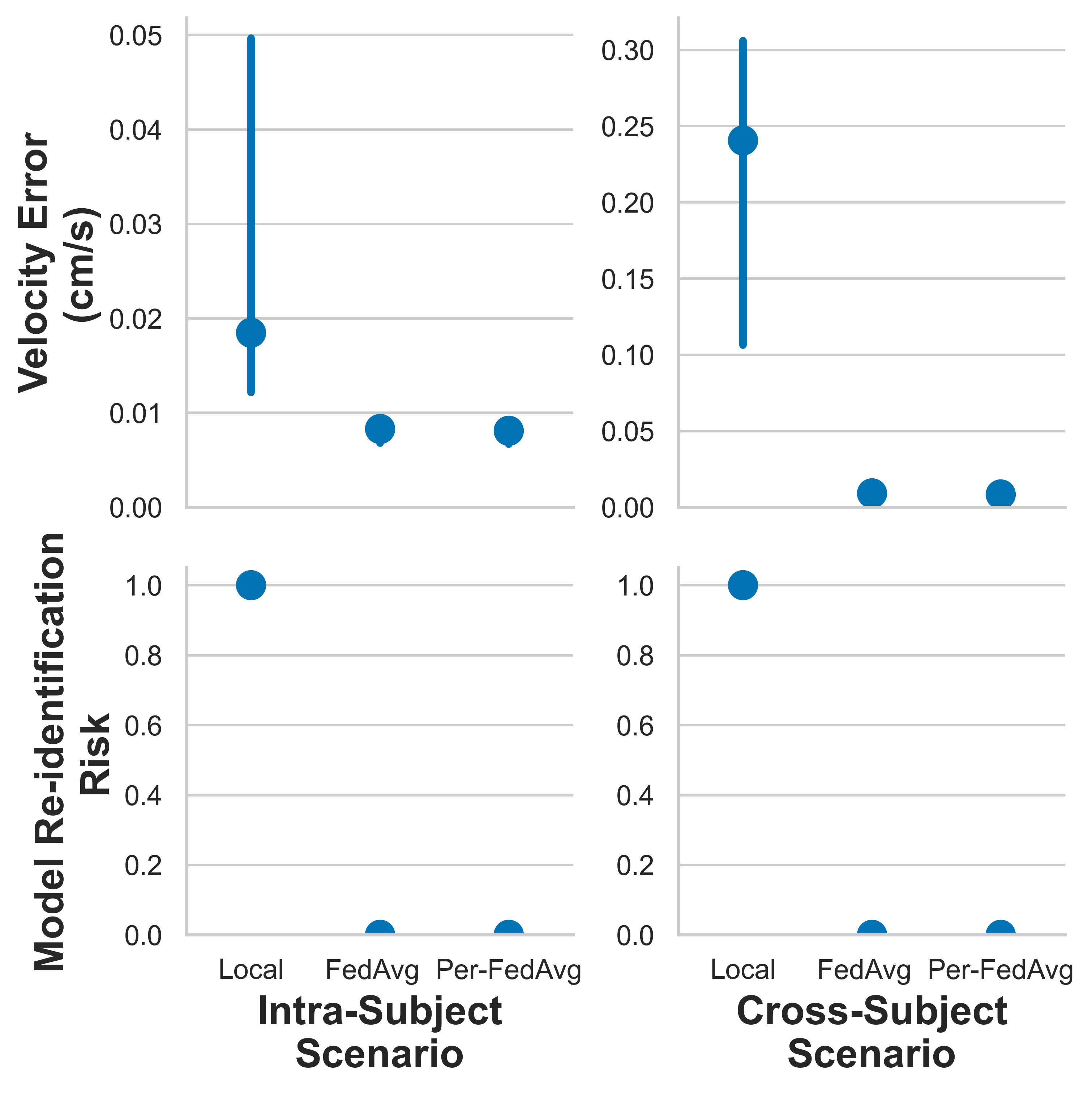}
  \caption{Offline performance (top) and privacy risk (bottom) for \CPHSAlgo{}, and Per-FedAvg \textit{Algorithms} ($N=14$ participants). 
  \textit{Scenarios} separated: Intra-subject (left) and Cross-subject (right). 
  For all plots, dots indicate the median and error bars represent the interquartile range (IQR: 25th, 75th percentile).
  For both metrics, lower values indicate better performance and improved privacy risk. 
  }
  %\Description{}
  \label{fig:OL_PrimaryOutcome}
\end{figure}

Both FL algorithms (FedAvg, Per-FedAvg) outperformed the \CPHSAlgo{} algorithm in Velocity Error in both the Intra-subject and Cross-subject scenarios. 
%Within each scenario, the two FL algorithms performed similarly: 
In the Intra-subject scenario, the difference between FedAvg and Per-FedAvg was not statistically significant ($p=0.27$), but there was a statistically significant difference in the Cross-subject scenario ($p<0.0001$). 
%For each algorithm, there was no statistically significant difference between the Intra- and Cross-subject scenarios results ($p>0.1$). 
% ^ I have no idea what this line is talking about. There claerly was a difference (that we even talk about later) for Velocity Error. There was none for Privacy but we didnt even run stats tests for OL Privacy...
% This was the original text there: "For all algorithms, there was no statistically significant difference between scenarios (p > 0.1)."
On average, FL algorithms outperformed \CPHSAlgo{} by 
\rev{0.011 cm/s [95\% CI: 0.004, 0.053] and 0.232 cm/s [95\% CI: 0.095, 0.298]}
in the Intra-subject and Cross-subject scenarios, respectively 
(\textit{Algorithm} main effect: $F(1.02,13.26)=49.8, p<0.0001$; Fig. \ref{fig:OL_PrimaryOutcome}, top). 
For all three algorithms, the Intra-subject Scenario had lower Velocity Errors than their Cross-subject counterparts (\textit{Scenario} main effect: $F(1,13)=5.1, p<0.05$; Fig. \ref{fig:OL_PrimaryOutcome}, top).
In both scenarios, \CPHSAlgo{} has both the highest average Velocity Error as well as the highest variance. 
There were no Velocity Error interaction effects ($F(1.01,13.20)=3.27, p=0.0928$), after Greenhouse-Geisser correction for sphericity violations.

For Model Re-identification Risk, all empirical results were either 0.0 (no privacy risk) or 1.0 (fully compromised privacy), with no differences between subjects within each condition. Thus, no statistical tests were performed as there was no within-group variation to test. 
\rev{We reiterate that the Model Re-identification Risk is the accuracy of our adversarial linkage classifier, and thus this metric is a continuous scale ranging from 0.0 to 1.0. In offline experimental results, the Local and FL conditions achieve the upper and lower bounds of this metric, respectively.}
It is clear that the algorithm drives the privacy risk, as \CPHSAlgo{} had a Model Re-identification Risk of 1.0 \rev{[IQR: 1.0, 1.0]}, whereas both FL algorithms had a Model Re-identification Risk of 0.0 \rev{[IQR: 0.0, 0.0]}.
Likewise, the global decoders of FedAvg and Per-FedAvg had a Model Re-identification Risk of 0.0 \rev{[IQR: 0.0, 0.0]} in both scenarios. 

\subsection{Online Result: Personalization-Privacy Trade-off} 

\begin{figure}[htbp]
  \centering
  \includegraphics[width=0.45\textwidth]{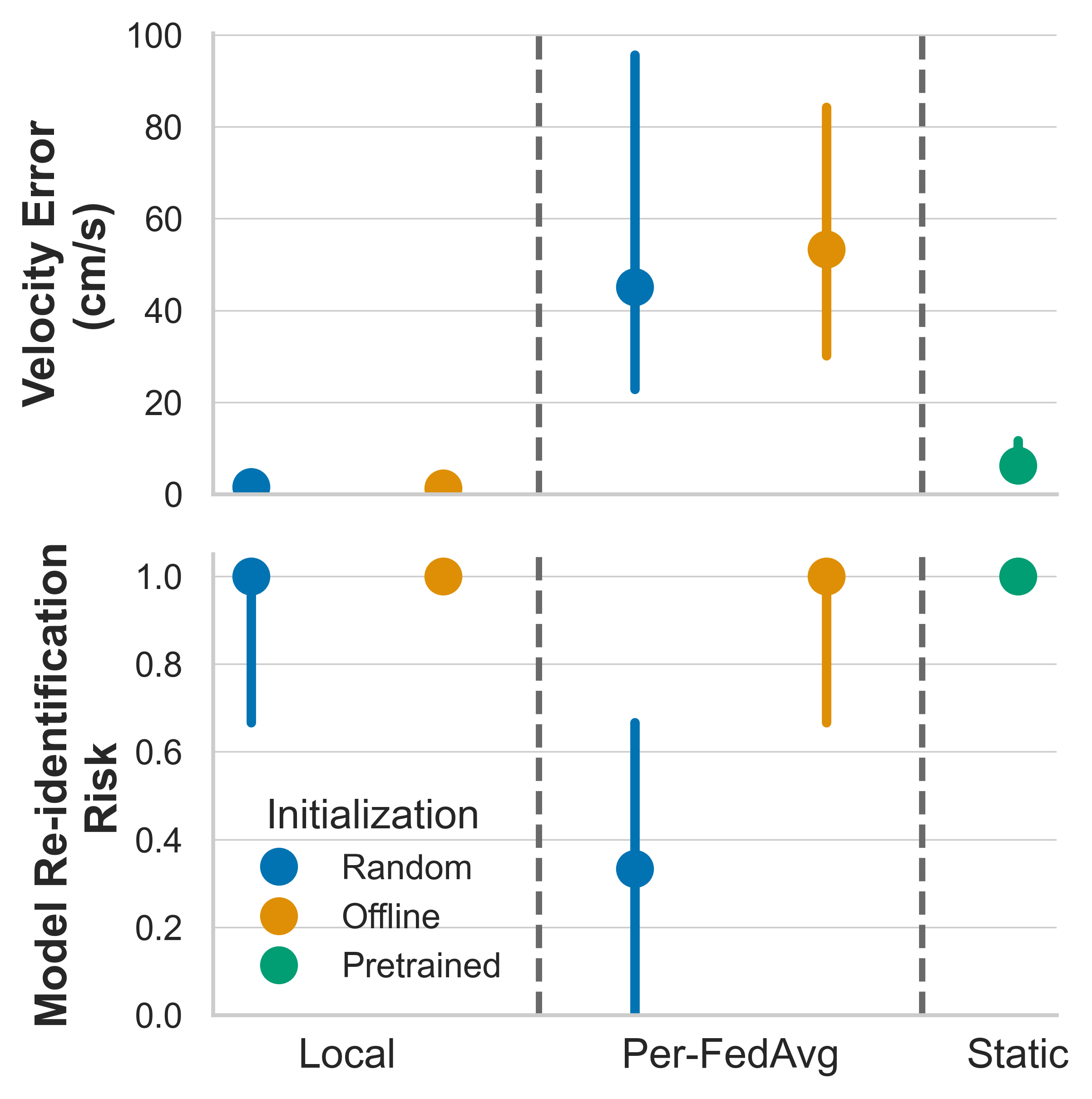}
  \caption{
  Experiment performance (Top) and privacy risk (bottom) for Local, Per-FedAvg, and Static \textit{Algorithms} with Random (light blue), offline (dark blue), and Pretrained (green) decoder \textit{Initializations} ($N=16$ participants). 
  For all plots, dots indicate the median and error bars represent the interquartile range (IQR: 25th, 75th percentile).
  For both metrics, lower values indicate better performance and improved privacy risk. 
  }
  \label{fig:CLE_PrimaryOutcomesVertical}
\end{figure}

The online experiments demonstrated a trade-off between performance and privacy, with \CPHSAlgo{} outperforming Per-FedAvg in Velocity Error by
\rev{43.7 cm/s [95\% CI: 20.3, 80.1] and 40.9 cm/s [95\% CI: 28.6, 65.7] for the Random and Offline Initializations, respectively}
(\textit{Algorithm} main effect: $F(2,60)=72.8, p<0.001$; Fig. \ref{fig:CLE_PrimaryOutcomesVertical}, top) 
but having a \rev{0.33 [95\% CI: 0.08, 0.50] (Random Init.) and 0.250 [95\% CI: 0.08, 0.50] (Offline Init.) higher Model Re-identification Risk} (\textit{Algorithm} main effect: $F(1,15)=23.0, p<0.001$; Fig. \ref{fig:CLE_PrimaryOutcomesVertical}, bottom).
The initialization did not have a statistically significant effect on Velocity Error (\textit{Initialization} main effect: $F(1,60)=1.2, p=0.268$; Fig. \ref{fig:CLE_PrimaryOutcomesVertical}, top).
Our planned post-hoc evaluations demonstrated that \CPHSAlgo{} outperformed Static by \rev{3.02 cm/s [95\% CI: 0.659, 9.260]
%; LSI: 3.212 [0.33 [95\% CI: 1.694, 9.624
} for Velocity Error ($p<0.05$; Fig. \ref{fig:CLE_PrimaryOutcomesVertical}, top).
There was no interaction effect for Velocity Error ($F(1,60)=1.0, p=0.308$).

For Model Re-identification Risk, our post-hoc evaluations showed that both \CPHSAlgo{} and Per-FedAvg improved privacy compared to Static by \rev{0.25 [95\% CI: 0.0, 0.33] (Random Init.); 0.0 [95\% CI: 0.0, 0.16] (Offline Init.)} ($p<0.05$; Fig. \ref{fig:CLE_PrimaryOutcomesVertical} bottom) and \rev{0.41 [95\% CI: 0.33, 0.67] (Random Init.); 0.33 [95\% CI: 0.17-0.50] (Offline Init.)} ($p<0.001$; Fig. \ref{fig:CLE_PrimaryOutcomesVertical} bottom), respectively.
Random initialization had a lower Model Re-identification Risk than the offline initialization by \rev{0.17 [95\% CI: 0, 0.33] (Local) and 0.17 [95\% CI: -0.08, 0.25] (Per-FedAvg)} (\textit{Initialization} main effect: $F(1,15)=5.2, p<0.05$; Fig. \ref{fig:CLE_PrimaryOutcomesVertical}, bottom).
No interaction effect was found for Model Re-identification Risk ($F(1,15)=0.02, p=0.88$; Fig. \ref{fig:CLE_PrimaryOutcomesVertical}, bottom).

\subsection{Offline and Online Decoder Convergence} 

\begin{figure}[htbp]
  \centering
  \includegraphics[width=0.5\textwidth]{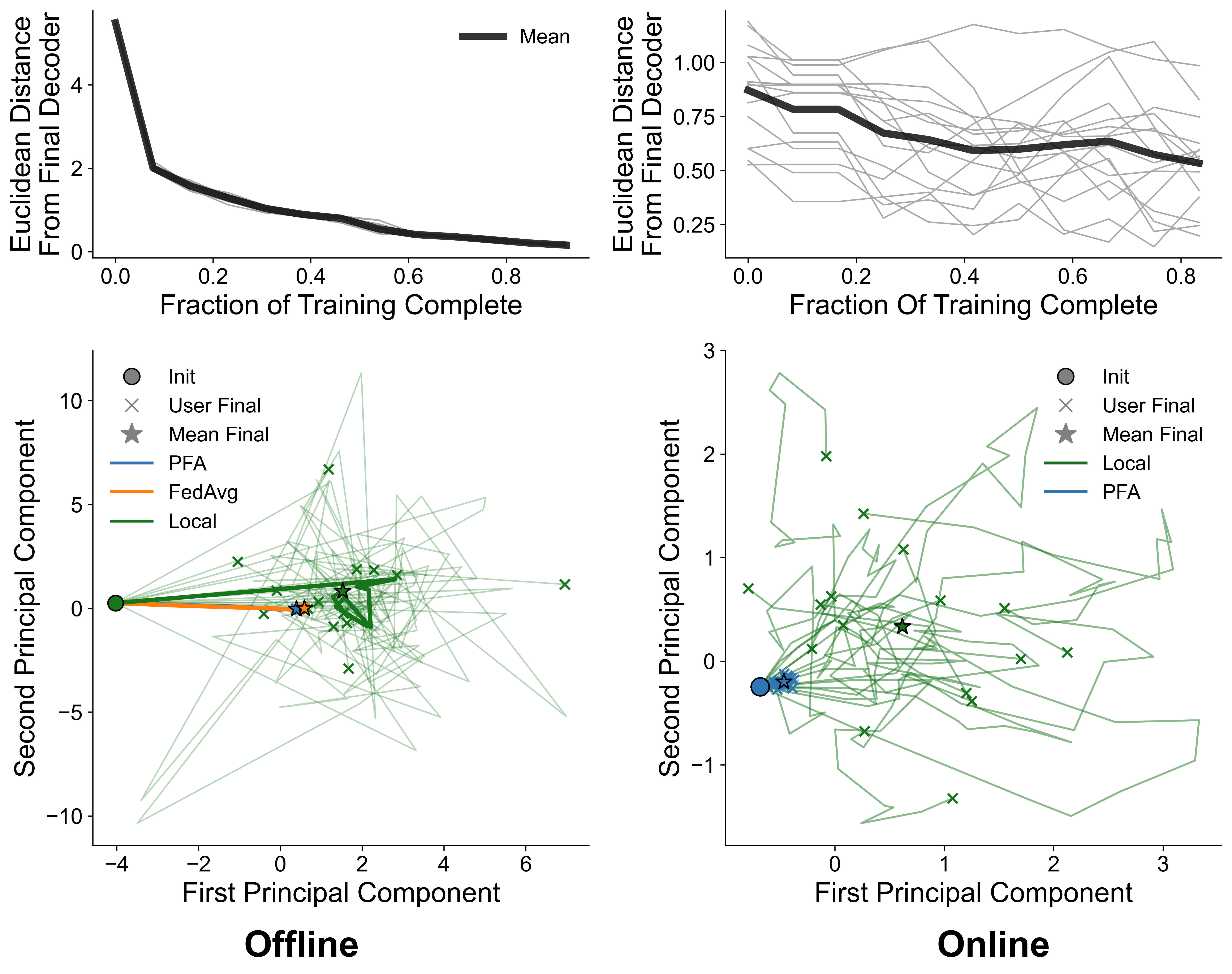}
  \caption{
  Top: Euclidean distance from the given update's decoder to the final decoder, for both offline (left) and online (right) evaluations.
  For the purposes of visualization and comparison, we solely plotted the Per-FedAvg (PFA) algorithm for both offline (Intra-subject scenario) and online (Random Initialization) evaluations. 
  Bottom: PCA applied to each update's decoder for all trials and plotted in a two-dimensional sub-space. Initial decoders are denoted with a circle marker, final decoders are denoted with x markers.
  The average final decoder, averaged across all users within each condition, is denoted with a star marker.
  For the purposes of clear visualization, only the Intra-subject scenario results are plotted for offline, and only Random Initialization results are plotted for online.
  The corresponding results (i.e., Cross-subject scenario results for offline, offline Initialization for online) followed nearly identical trajectories. 
  The full plot can be found in the Supplementary Materials.
  }
  \label{fig:DecAn}
\end{figure}

We analyzed decoder changes during training between offline and online evaluations (Fig. \ref{fig:DecAn}, top).
For the offline evaluation, the decoders converged smoothly to the final decoder parameters in an exponential-like decay pattern for all subjects (Fig. \ref{fig:DecAn}, top left).
In contrast, online decoder distances exhibited fluctuations with only a weak downward trend (Fig.~\ref{fig:DecAn}, top right).
These trends were also reflected in PCA projections of decoder trajectories in a two-dimensional subspace (Fig.\ref{fig:DecAn}, bottom). 
In the offline evaluation, the FedAvg and Per-FedAvg decoders for all users converge to the same points in both Intra-subject and Cross-subject scenarios (Fig. \ref{fig:DecAn}, bottom left).
In online experiments, FL decoders exhibited short trajectories, with final decoders tightly clustered near their initialization. 
In contrast, \CPHSAlgo{} decoders did not converge to a common solution across users (Fig.~\ref{fig:DecAn}, bottom right). 
In both offline and online settings, \CPHSAlgo{} trajectories are dispersed and variable in the sub-space, while Per-FedAvg trajectories are more tightly clustered and closer to the initial decoders.

\section{Discussion}

\rev{The goal of this study was to investigate the generalizability of FL approaches from offline simulation to an online neural interface. We establish a framework for selecting appropriate privacy-preserving FL algorithms depending on the experimental neural interface scenario. We demonstrated a distinct performance-privacy tradeoff in the application of FL to neural interfaces. In offline simulations, FL achieved a significant reduction in decoding error, improving tracking precision by up to 22\%, while simultaneously nullifying the adversarial linkage attack. Notably, Local decoders were 100\% identifiable by the adversary. Conversely, real-time online experiments showed that these offline results did not generalize to the online setting. In online, Local decoders outperformed the modified Per-FedAvg approach by 47.7 cm/s on average. Nonetheless, the modified Per-FedAvg still offered greater privacy protection than its local learning counterpart. These results establish that although FL is highly effective in offline datasets, straightforward adaptations struggle to generalize to sequential-user environments.}

\subsection{Consistent Benefits in Offline Federated Learning}

Federated learning demonstrated clear advantages over local learning approaches in our offline simulations, consistent with prior applications of FL to offline neural interfaces \cite{OL_FL_EEG_Emotion, OL_FL_BCI, OL_FL_RT_GestureRecog, OL_FL_RT_HRI}.
In EEG-based BCIs, Liu et al.~\cite{OL_FL_BCI} reported performance improvements of up to 8.4\% when aggregating across users via FL, and Xu et al.~\cite{OL_FL_EEG_Emotion} observed a 29.3\% drop in accuracy when switching from FL to local learning for emotion recognition.
Likewise, Zafar et al. \cite{OL_FL_RT_GestureRecog} demonstrated success with FedAvg for EMG gesture recognition, finding that FedAvg outperformed FedSGD, FedProx, and a non-federated deep learning baseline.
These findings highlight FL’s ability to improve generalization by training over multiple users without requiring direct access to raw neural data. 

Our offline results align with this pattern: both FedAvg and Per-FedAvg achieved substantially lower Velocity Error than local learning in both intra- and cross-subject scenarios.
A possible explanation is that multi-user aggregation implicitly smooths the optimization landscape.
Theoretical work on stochastic optimization has shown that larger effective batch sizes reduce gradient variance \cite{nesterov2018lectures}, yielding more stable and reliable gradient updates.
Although we did not directly measure gradient noise, the smoother convergence trajectories exhibited by FL decoders in our offline experiments, compared to the higher-variance trajectories of \CPHSAlgo{}, are consistent with this interpretation.
We offer this as a potential mechanism rather than a definitive causal claim.

% Explanation/context of results
\subsection{Challenges with Online Federated Learning}

Although FL provided clear performance benefits in our offline simulations, these advantages did not translate directly to the online setting. 
In real-time experiments, the \CPHSAlgo{} outperformed the modified Per-FedAvg algorithm, although the latter continued to provide strong privacy protection.
We note that \CPHSAlgo{} performed in line with previous works \cite{CPHS} in terms of Tracking (i.e., Position) Error, as seen in Supplemental Figure 3.
This divergence between offline and online FL performance highlights fundamental challenges in applying FL to sequential-user neural decoding.

The primary challenge in translating FL benefits to real-time applications is the mismatch between theoretical assumptions and the practical structure of training neural decoders. 
Standard federated algorithms rely on the simultaneous availability of multiple users to provide inter-user regularization. 
In our online experiments (i.e., our designated \textit{sequential} scenario), simultaneous aggregation was not possible without storing raw user data, thereby undermining the privacy goals of FL.
Our modified approach instead regularized the model against a global initialization, which, while necessary for the real-time user study, lacks the collaborative learning potential of concurrent multi-user updates.

Furthermore, online neural decoding is uniquely sensitive to hyperparameter selection.
Several hyperparameters, including the number of gradient steps per update, batch size, penalty terms, and learning rate, were critical for ensuring stable co-adaptive behavior,  convergence, and real-time latency. 
Hyperparameters that yielded smooth convergence in offline training were often unstable in online experiments, resulting in erratic cursor movements or user frustration. 
These observations align with prior reports that closed-loop decoder adaptation requires extensive, and in many cases personalized, tuning and interaction before the interface is stable enough to learn~\cite{Amy_CLDA_SkillfulNeuroProsCtrl, Amy_DesignHyperparams}.
The limited control observed in our online federated trials suggests that the tethering of a local decoder to a global model may impede the user's ability to effectively partner with the algorithm. 
This underscores the need for adaptive federated approaches that can dynamically balance global regularization with the high degree of personalization required for stable, real-time neural control.

\subsection{Generalizability of Privacy Risks Across FL Architectures}

\rev{
The scope of the privacy risks identified in this study extends beyond the specific implementation of linear decoders and weight-based aggregation. In this study, we employed weight-based aggregation, a standard protocol in FL that is functionally equivalent to gradient-based updates in its optimization goals. This approach has been validated across a range of architectures, from generalized linear models \cite{FL_LinearModels, FL_GLM_EHR, FL_GLM_PP} to deep neural networks \cite{FedAvg, Per-FedAvg, OL_FL_BCI}. Although this work employs linear decoders, the observed re-identification risk is likely to generalize to more complex, high-capacity architectures (i.e., deep learning) because the exploited vulnerability is fundamentally rooted in cross-user physiological heterogeneity, rather than being a specific property of linear models or the weight-based averaging scheme. In fact, as model capacity increases, the potential for these models to encode more granular, subject-specific signatures likely intensifies the threat of re-identification. Consequently, our findings provide a conservative lower bound for the privacy risks inherent in high-dimensional neural interfaces.}

\subsection{Limitations and Future Directions}
\rev{This study was conducted in a well-controlled environment, leveraging high-density, high signal-to-noise ratio (SNR) data and linear models. Although these conditions represent an ideal scenario, this was a deliberate methodological baseline intended to isolate the effects of federated optimization from confounding variables such as complex model architectures or real-world signal noise. Establishing this baseline is a critical first step before extending FL frameworks for neural decoding to motor-impaired populations or unconstrained environments.

It is important to note that the findings presented here likely represent a conservative lower bound for the challenges of online FL. In more complex, noisier scenarios, the trade-offs between local adaptation and global aggregation are expected to become even more pronounced. Furthermore, although we used linear decoders, the identified re-identification risks stem from the inherent heterogeneity of physiological data. The use of higher-capacity models, such as deep neural networks, would likely increase the risk of model weights encoding unique subject fingerprints rather than mitigating these privacy concerns. Future work should build on this foundation by evaluating these federated architectures in individuals with motor impairments and by employing nonlinear models for continuous, multi-degree-of-freedom control.
}

Our framework provides practical guidance for selecting FL algorithms based on personalized cost and user availability. 
However, the framework is inherently limited in prescriptiveness, as it does not specify implementation details or the ``best'' algorithm. 
This is because there is no FL algorithm that dominates all tasks across all domains, and outside of task-specific benchmarks (e.g., CIFAR-100 \cite{CIFAR100} for image classification), researchers cannot know how a given FL algorithm will extend to their specific domain and task. 
The lack of standardized benchmarks for neural decoding remains a hurdle for the field, necessitating extensive domain-specific tuning for each new application.
For implementation-oriented tools, we point to Flower~\cite{flower}, which provides the infrastructure for setting up a distributed learning scenario, but minimal guidance on selecting the FL algorithm (particularly within the context of online neural interfaces).
Future work should develop FL algorithms explicitly tailored for sequential, real-time training scenarios, as well as adaptive hyperparameter strategies suited for online, co-adaptive environments.

%Future work should investigate how different model capacities influence the trade-off between decoding performance and user privacy.

% \kai{Say something about why Local and FL are not directly comparable due to first and second order differences?}

%\hl{is there a smart way to train the decoder in online? eg bring BCI efficient people in first and then bring in BCI inefficient people. }
%This challenge is compounded by user variability: tuning hyperparameters on a single subject is insufficient when working with heterogeneous neural signals, and even pilot studies may not yield hyperparameters that generalize to new users. 
%online conditions severely limit feasible hyperparameter ranges, update frequencies, and training step counts, factors easily managed in simulation but problematic in interactive studies. 

%%% EXPLANATION FOR THE DISCREPANCY BETWEEN OL LOCAL AND CL LOCAL
%We note that our approach used the ridge regression cost function with a linear regression model, and thus has a convex loss landscape. Thus, in theory, each subject has a single global minimum, which can be solved in closed-form (only after all the data has been collected). It is not necessarily the case that each subject has the same global minimum, however, from Figure \ref{fig:DecAn} lower left, it appears that all subjects in the offline simulations had the same or at least very similar global minima, at least in this 2D latent representation.

\section{Conclusion}
This study offers the first empirical exploration of federated learning in both offline and online neural decoding contexts. By proposing a conceptual framework centered on personalized cost and user availability, we provide a roadmap for navigating the distinct constraints of neural interface applications.

Our findings confirm that while FL provides a powerful synergy of performance and privacy in static, offline settings, this relationship shifts significantly in real-time environments. The performance gap between FL and local decoders observed in our online user study reveals that the dynamics of online decoder adaptation and strict real-world experimental constraints introduce unique challenges for existing federated optimization techniques.

Ultimately, this work demonstrates that privacy-preserving neural interfaces are feasible but require a departure from standard, simulation-based FL assumptions. Our results underscore the need for a new class of adaptive federated algorithms specifically optimized for the stability and personalization requirements of real-time neural control.

% \section*{References}
\bibliography{references.bib}

\end{document}

\section{Supplementary Material}

\begin{figure}[htbp]
  \centering
  \includegraphics[width=0.85\textwidth]{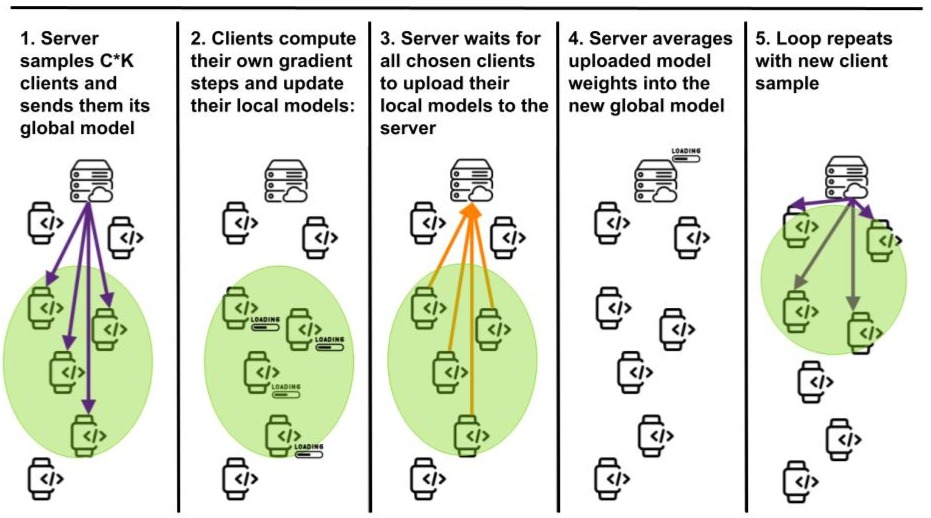}
  \caption{
  Depiction of a single federated training round in FedAvg [47].
  }
  \label{fig:CLE_TrackingError}
\end{figure}

\begin{figure}[htbp]
  \centering
  \includegraphics[width=0.9\textwidth]{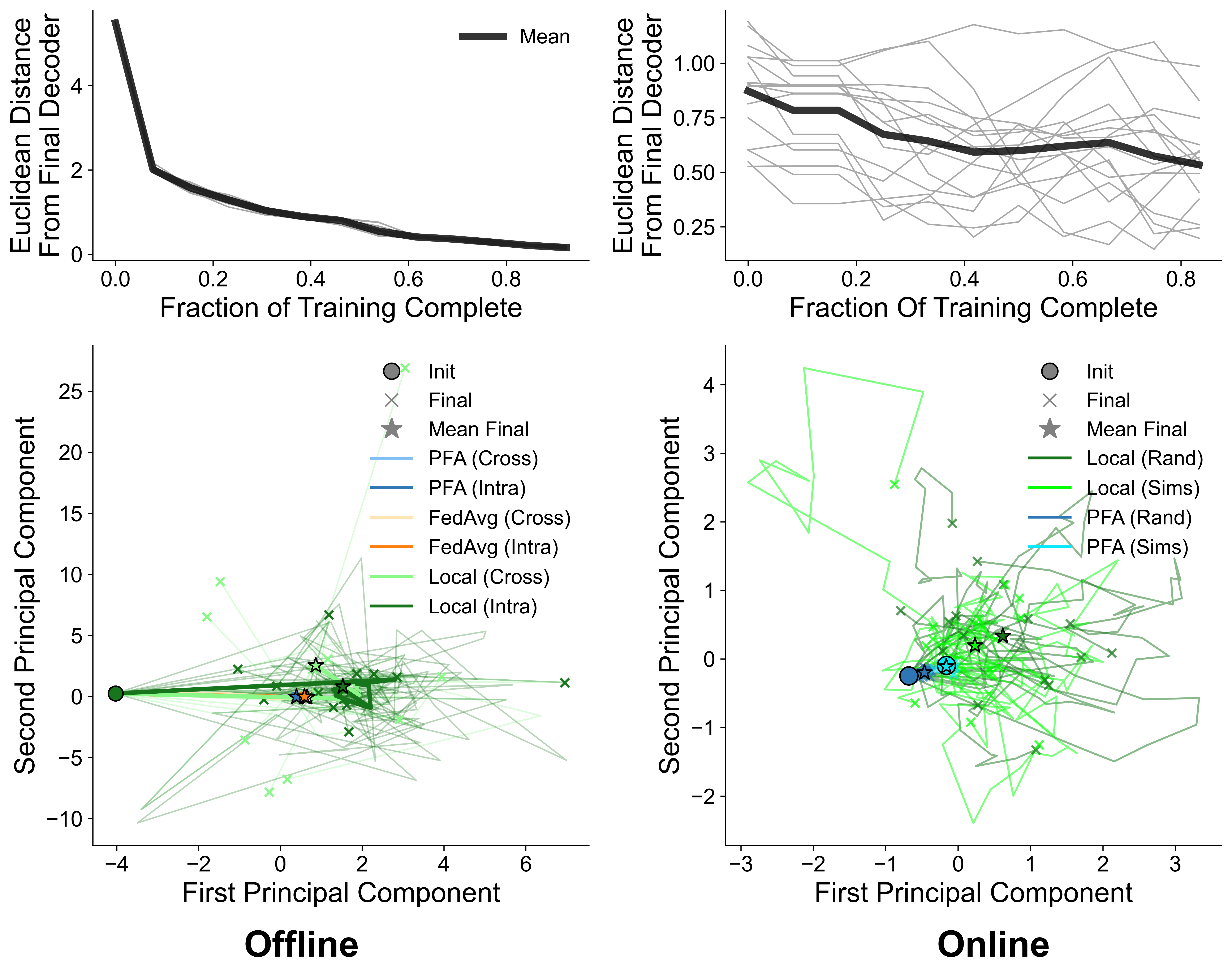}
  \caption{
  Top: Euclidean distance from the current update's decoder to the final decoder, for both offline (left) and online (right) evaluations.
  For the purposes of visualization and comparison, we solely plotted the Per-FedAvg algorithm for both offline (intra-subject scenario) and online (random initialization) evaluations. 
  Bottom: PCA applied to flattened decoders and plotted in a two-dimensional sub-space. Initial decoders are denoted with a circle marker, final decoders are denoted with x markers.
  The final decoder, averaged across all users for each condition, is denoted with a star marker.
  }
  \label{fig:DecAn_Appendix}
\end{figure}

\begin{figure}[htbp]
  \centering
  \includegraphics[width=0.65\textwidth]{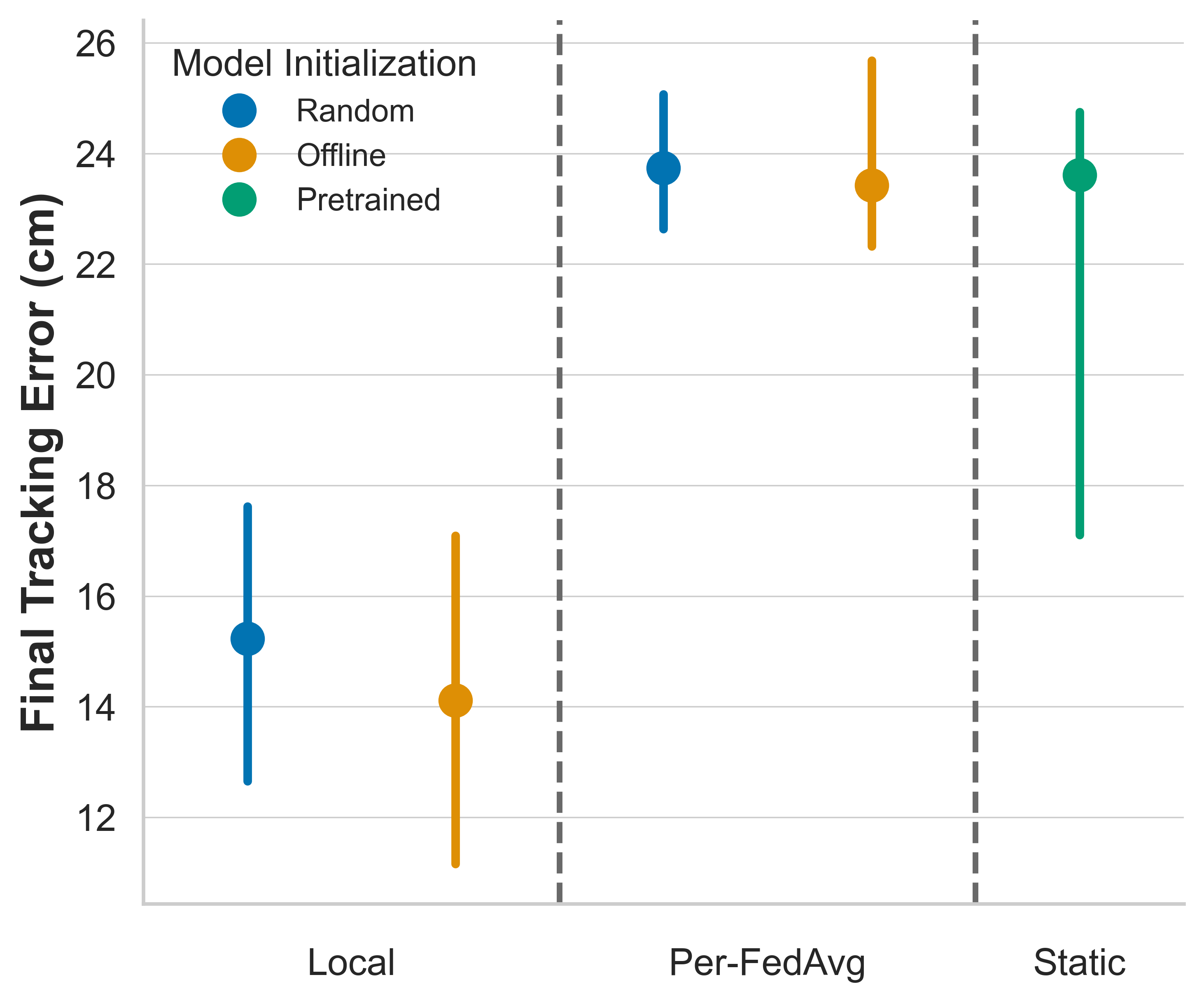}
  \caption{
  Tracking (position) error from the final full update in the online experiment.
  }
  \label{fig:CLE_TrackingError}
\end{figure}